\documentclass[conference]{IEEEtran}
\IEEEoverridecommandlockouts
\usepackage{cite}
\usepackage{amsmath,amssymb,amsfonts}

\usepackage{algorithm}
\usepackage{algpseudocode}
\usepackage{graphicx}
\usepackage{textcomp}
\usepackage{xcolor}
\usepackage{multirow}

\def\BibTeX{{\rm B\kern-.05em{\sc i\kern-.025em b}\kern-.08em
    T\kern-.1667em\lower.7ex\hbox{E}\kern-.125emX}}
\begin{document}

\title{Simultaneous Adversarial Attacks On Multiple Face Recognition System Components}

\author{\IEEEauthorblockN{Inderjeet Singh, Kazuya Kakizaki, Toshinori Araki}
\IEEEauthorblockA{\textit{Secure System Platform Research Laboratories} \\
\textit{NEC Corporation}\\
7-1, Shiba 5-chome Minato-ku, Tokyo 108-8001 Japan \\
\texttt{\{inderjeet78, kazuya1210, toshinori\_araki\}@nec.com}}}

\maketitle

\begin{abstract}
In this work, we investigate the potential threat of adversarial examples to the security of face recognition systems. Although previous research has explored the adversarial risk to individual components of FRSs, our study presents an initial exploration of an adversary simultaneously fooling multiple components: the face detector and feature extractor in an FRS pipeline. We propose three multi-objective attacks on FRSs and demonstrate their effectiveness through a preliminary experimental analysis on a target system. Our attacks achieved up to 100\% Attack Success Rates against both the face detector and feature extractor and were able to manipulate the face detection probability by up to 50\% depending on the adversarial objective. This research identifies and examines novel attack vectors against FRSs and suggests possible ways to augment the robustness by leveraging the attack vector's knowledge during training of an FRS's components.
\end{abstract}

\begin{IEEEkeywords}
adversarial example, face recognition, risk assessment, robustness
\end{IEEEkeywords}

\section{Introduction}
The increasing use of face recognition systems (FRSs) in various domains, such as security, law enforcement, and customer identification, has brought attention to the robustness and trustworthiness of these systems when faced with malicious inputs, known as adversarial examples (AXs)\cite{sharif2019general}\cite{guetta2021dodging}\cite{singh2022powerful}. AXs are instances with carefully crafted noise \cite{pgd}, inserted by adversaries with the aim of fooling a target deep learning (DL) model. The practical pipelines of FRSs consist of two primary DL components: a face detector for detecting the presence and location of a face, and a face matcher, that compares the match on features extracted by a DL model.

\begin{figure}[t]
\centerline{\includegraphics[width=0.95\linewidth]{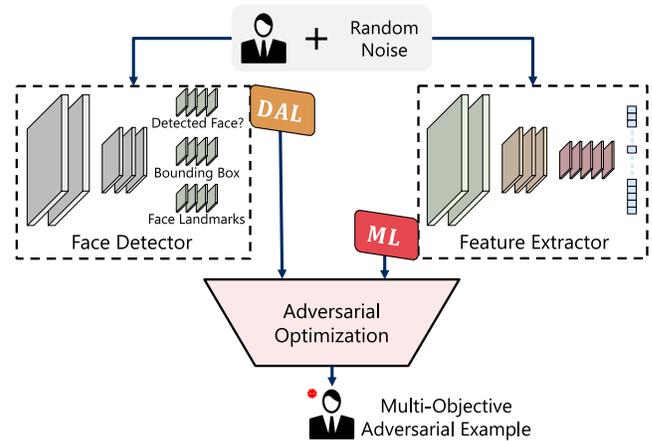}}
\caption{A visual representation of our proposed multi-objective AX generation on the face recognition pipeline's face detection and feature extractor-based matching stages is shown. An initial perturbation is introduced to a clean facial image, which is then presented to a face detection model and a feature extraction model. Subsequent to this, an adversarial optimization process combining the detection adversarial loss (DAL) and misclassification loss (ML) is conducted to transform the perturbation into an adversarial noise, resulting in a multi-objective AX.}
\label{fig1}
\end{figure}

Previous research has extensively studied the susceptibility of DL models in FRSs to AXs and has often only considered the adversarial risk to a single component, rather than evaluating the system as a whole. This research endeavors to identify and examine novel methods for attacking FRSs. We propose adversarial attacks that can be tailored to a variety of objectives and can effectively deceive \textit{multiple} components within an FRS. To demonstrate this concept, we present the creation of AXs that can deceive both the face detection and face matching modules having significantly disparate architectures and training objectives within an FRS, as depicted in Fig. \ref{fig1}.

We propose three different multi-objective attacks on FRSs. The first is an impersonation attack that performs impersonation on the face matcher while still being detectable by the face detector. The second attack is an evasion attack that is able to evade classification by the face matcher while still being detectable by the face detector. The third attack is also an evasion attack, but it is able to evade both the face detector and matcher. Through a comprehensive experimental analysis, we demonstrate the effectiveness of our proposed attacks on a target FRS that utilizes a multi-stage face detector and a Squeeze and Excitation Inception ResNet model as a face feature extractor for the face matcher.

Our multi-objective evasion and impersonation attacks achieved up to 100\% Attack Success Rates (ASRs) against both the face detector and the face matcher. Our attacks against the face detector resulted in a change of up to 50\% in the face detection probability, depending on the specific adversarial objective used. The proposed attacks facilitate a thorough evaluation of the vulnerabilities of practical FRSs to sophisticated adversarial attacks and can be leveraged to augment the robustness of face detection and feature extraction models through techniques like adversarial training.

\section{Preliminaries}
\subsection{Face Recognition System}
FRSs leverage the power of DL models to automatically learn complex features from face images and perform detection and identification tasks with high accuracy. These systems generally adhere to a pipeline that encompasses multiple interconnected stages, including input capture via a camera, pre-processing of the captured input to make it suitable for the detection model, face detection using a \textit{face detector} \cite{zhang2016joint}, post-processing with cropping of identified faces in the images to prepare them for the face feature extractor, DL-based face feature extraction \cite{kakizaki2021toward} and matching which is also called \textit{face matcher} in the FRS pipeline.

\subsubsection{Face Detector}
A face detector locates faces in digital images or video frames by identifying patterns that are characteristic of faces, such as the presence of eyes, nose, and mouth. Early face detectors used hand-crafted features for the face detection. Modern face detectors are DL-based and they use single stage or multi-stage detection steps to output refined predictions about the presence and locations of the faces in the image. In this work, we consider a well-known multi-stage face detector, \textit{Multi-Task Cascaded Convolutional Neural Network (MTCNN)} \cite{zhang2016joint} because of its complex architecture making it challenging for adversary to attack.

MTCNN is a face detection and alignment algorithm that uses cascaded convolutional neural networks (CNNs). It consists of three interconnected CNNs, known as the \textit{Proposal Network (P-Net)}, the \textit{Refine Network (R-Net)}, and the \textit{Output Network (O-Net)}. The P-Net generates candidate face bounding boxes, which are refined by the R-Net. The O-Net then further refine and detects five facial landmarks within each refined bounding box. MTCNN is highly accurate and efficient, and has been widely used in various applications.

\subsubsection{Face Matcher}
The Face Matcher module in FRSs encompasses a Face feature extractor and a metric to compare the similarities among different faces. The feature extractor is based on deep metric learning (DML). It is trained using similarity or distance-based DML loss functions to extract discriminative features from the faces of different persons. Finally, the extracted features by the feature extractor are compared using a distance or similarity metric against the features of different targets or registered faces to make the classification and authorization of the input.

\subsection{Adversarial Attacks Against Deep Learning Models}
Adversarial attacks are instances of presenting carefully crafted inputs, known as AXs, to a DL model to achieve an adversary's desired prediction outcome. An AX $x_{adv}$, which is a clean sample $x$ added  with adversarial noise $\delta$ is crafted following the objective as
\begin{equation}
\min_{\delta\in \mathbb{S}} \mathcal{L}\left(\theta, (x+\delta), y_{adv}\right),
\end{equation}

\noindent where $y_{adv}$ is the adversary's target label depending on the type of attack, $\theta$ are the DL model's parameters, $\mathbb{S}$ is feasible solution space for the adversarial noise, $\mathcal{L}$ is the function for calculating the adversarial loss. 

AXs may include imperceptibly small or physically imperceptible perturbations, and can result in evasion or impersonation against the model.

\subsubsection{Evasion Attacks}
Crafting an AX with adversarial noise $\delta_{eva}$ to avoid true classification while optimizing:
\begin{equation}
\max_{{\delta_{eva}}\in \mathbb{S}} \mathcal{L}\left(\theta, (x+{\delta_{eva}}), y_{true}\right).
\end{equation}

\subsubsection{Impersonation Attacks}
Adversary crafts an adversarial noise $\delta_{imper}$ to make the target model classify the input instance in the desired class $y_{t}\neq y_{true}$ while optimizing:
\begin{equation}
\min_{{\delta_{imper}}\in \mathbb{S}} \mathcal{L}\left(\theta, (x+{\delta_{imper}}), y_{true}\right).
\end{equation}

\section{Related Works}

Recent advancements in the field of adversarial attacks have brought to light the vulnerability of FRSs to such attacks, owing to their widespread deployment in various applications. Despite this, prior research has primarily concentrated on the targeting of specific components within FRSs, such as the face detection or facial matching module. 

\noindent \textbf{Attacks Against Face Detectors: }
Adversarial attacks targeting \textit{face detection} systems generally aim to modify or conceal detectable facial features in order to evade detection. Yang et al. \cite{yang2020closer} designed universal adversarial face-like patches that can effectively evade face detection systems by generating perturbations that are imperceptible to humans. Hoory et al. \cite{hoory2020dynamic} proposed a dynamic adversarial patch that uses a screen that can switch the patch depending on the camera's position, allowing the patch to evade face detection models even when the camera angle changes. This approach is an improvement over previous works, which often relied on fixed patches that could only evade face detection models from certain angles. Thys et al. \cite{thys2019fooling} developed an adversarial patch that is capable of fooling person detection systems by generating perturbations that are not noticeable to humans, but can significantly degrade the performance of the person detection model. Kaziakhmedov et al. \cite{kaziakhmedov2019real} also proposed an adversarial patch attack that is robust against brightness changes and can effectively evade the well-known multi-stage MTCNN face detector \cite{zhang2016joint}. However, these attacks against face detectors do not consider fooling the face matcher, which is an essential component of FRSs. As a result, these attacks are inadequate for assessing the real-world risk of face recognition systems.

\noindent \textbf{Attacks Against Face Matchers: }
Previous research on such attacks has focused on adding noise or perturbations to the input data, such as altering pixel values \cite{dabouei2019fast}, adding features like glasses \cite{sharif2019general}\cite{singh2021brightness}\cite{singh2022powerful}, hats \cite{komkov2021advhat} or object patches \cite{wei2022adversarial},  hair color \cite{kakizaki2019adversarial}, introducing significant pixel and grid-based noise \cite{goswami2018unravelling}\cite{guetta2021dodging}, visible light \cite{shen2019vla}, and imperceptible face noise \cite{chatzikyriakidis2019adversarial}\cite{kakizaki2021toward}. These perturbations have been demonstrated to significantly impact the accuracy of FRSs, but they only consider fooling the \textit{face matcher} in the FRS.

In summary, previous research has not addressed the creation of multi-objective AXs that can simultaneously target both face detectors and face matchers. These multi-objective AXs offer significantly higher ASR guarantees for the attacker to fool a target FRS as a whole, thereby posing a significant practical threat. Our paper presents a novel approach for generating such attacks, which can be used to effectively assess the risk of practical FRSs and to train FRSs to be more resilient to these types of attacks through adversarial training techniques \cite{advtraining}. This represents a significant advancement in the field of adversarial machine learning for FRSs.

\section{Method}
In this work, we present three novel attacks which simultaneously deceive face detectors and feature extractors: (1) the \textit{Detectable-Impersonation Attack (DI-Attack)}, which generates a highly effective impersonation attack while still allowing the attacker's face to be detected by the target FRS's face detector during the attack process; (2) the \textit{Detectable-Evasion Attack (DE-Attack)}, which enables evasion against the face feature extraction and matching phase while still remaining detectable to the face detector; and (3) the \textit{Undetectable-Evasion Attack (UE-Attack)}, the most advanced evasion attack against FRSs, ensuring evasion against both face detection and face feature extraction and matching in the target FRS. An outline of our methods is presented in Algorithm \ref{alg:attack_gen_procedure}.

\begin{algorithm}
\caption{Multi-Objective Attack Generation Procedure}\label{alg:attack_gen_procedure}
\begin{algorithmic}
\State \hskip-0.6em \textbf{Inputs:} Source image $x_s$, target image $x_t$; random noise $\delta$; Adversarial losses for detection $\mathcal{L}_{det}$ and misclassification $\mathcal{L}_{mis}$; Loss weight parameter $\alpha$; attack method $PGD$; target face detector $d$ and feature extractor $f$;
\State \hskip-0.6em \textbf{While} $PGD$ not finished \textbf{do}
\State Add random noise $\delta$ to source image $x_s$, i.e., $x = x_s + \delta$;
\State Calculate detection adversarial loss $\mathcal{L}_{det}(d,x)$;
\State Calculate misclassification loss $\mathcal{L}_{mis}(f, x, x_t)$;
\State Perform white-box PGD update $\delta \leftarrow PGD(f, d, \alpha\cdot\mathcal{L}_{det}+\mathcal{L}_{mis}, x)$;

\State \hskip-0.6em \textbf{Return} Multi-Objective AX, $(x_s+\delta)$;
\end{algorithmic}
\end{algorithm}

\subsection{DI-Attack}
The DI-Attack (Detectable-Impersonation Attack) method involves crafting impersonation AXs to deceive the face matcher in a FRS while still being detectable by the face detector.

\noindent\textbf{Threat Model.} The size of the adversarial noise or patch is a crucial factor in determining the success of an attack. \textit{Larger} adversarial noise sizes offer several benefits to the attacker, including a higher probability of attack success, increased attack robustness, and faster convergence, which can lead to lower time and computational requirements. However, there is a \textit{trade-off} to consider: face images with large adversarial noise may be less detectable to the face detector in the target FRS, causing impersonation attacks to fail. For an impersonation attack to be successful, the adversarial input must not be discarded at any prior stage in the target FRS and must reach the face matching stage. Therefore, it is important to carefully optimize the adversarial noise to maximize attack success while minimizing the risk of failed detection.

To balance these competing factors, it is important for the adversary to carefully optimize the adversarial noise to satisfy the detectability constraint. This requires a nuanced understanding of the relationship between patch size, attack success probability, and detectability. By carefully considering these variables, we show that it is possible to craft powerful and robust impersonation attacks with \textit{large adversarial patch sizes} that remain detectable to the face detection while successfully executing impersonation attacks against the face matcher of the target FRS.

\noindent\textbf{Attack fabrication. }
We craft attacks using white-box projected gradient descent (PGD) method \cite{pgd} to fool both face detector and matcher during inference.  

\subsubsection{Attacking MTCNN}
We craft adversarial noise to ensure that the detection probability of the face inside input image do not gets reduced even after applying \textit{large} adversarial noise. Since MTCNN has multiple CNNs for each stage, we demonstrate that our method allows crafting robust attacks by only accessing the \textit{PNet} of the MTCNN. We define an adversarial loss $\mathcal{L}_{det-D}$ against the PNet to falsely increase the detection probability of the face in the input image with adversarial noise.

The detection loss $\mathcal{L}_{det-D}$ in the $i^{th}$ iteration of attack generation process is calculated by penalizing the reduction in the probability of a face due to addition of a patch, i.e. 
\begin{equation}\label{det-D loss}
    \mathcal{L}_{det-D}^{i} = \sum_{k \in 1,2..m} \sum_{l \in 1,2..n} (K \cdot A_{k,l}- A_{k,l}\cdot Y_{k,l}^{i})^{s},
\end{equation}

\noindent where $Y_{k,l}^{i}$ is the probability of presence of a face in PNet's $12\times12$ window at pixel location $(k,l)$ in the input image. $A_{k, l}= \begin{cases}1 \text { if } & T_{k, l} \geq \delta \\ 0 & \text { else }\end{cases}$; $T_{k, l}$ is the probability of presence of a face at $(k,l)$ in the source image and $\delta$ is the predefined detection threshold for the PNet. $m=\left\lfloor\frac{W-12}{4}\right\rfloor+1$ and $n=\left\lfloor\frac{H-12}{4}\right\rfloor+1$; $W$ and $H$ are the width and height of the input image to the PNet of the MTCNN. $s>0$ is real number responsible for exponentiality of the loss landscape and $K$ is a margin parameter.

\subsubsection{Attacking Face Matcher}
After optimizing the attack procedure to satisfy the detectability constraints, adversarial objective against the facial feature extraction stage is developed to satisfy the impersonation attack objective. We define an impersonation loss $\mathcal{L}_{imper}$ against the trained feature extractor $f$ of the target FRS.

The impersonation loss $\mathcal{L}_{imper}$ in the $i^{th}$ iteration of attack generation process is defined as
\begin{equation}
\mathcal{L}_{\text {imper }}^i=\left.\left\|f(t)-f\left(s^i\right)\right\|\right|_p.
\end{equation}

\subsubsection{Calculating Total Loss}
Finally we combine the detection loss $\mathcal{L}_{det-D}$ and the impersonation loss $\mathcal{L}_{imper}$, and calculate the total loss $\mathcal{L}_{\text {total }}^i$ in $i^{th}$ training iteration as
\begin{equation}
\mathcal{L}_{\text {total }}^i=\alpha \cdot \mathcal{L}_{\text {det-D }}^i+\mathcal{L}_{\text {imper }}^i,
\end{equation}

\noindent where $\alpha$ is a weight parameter for the detection loss.

The adversarial noise is updated using the PGD attack method until it simultaneously satisfies the detection and impersonation objectives. 

\subsection{DE-Attack}
The DE-Attack (Detectable-Evasion Attack) method involves crafting evasion AXs to evade the face matcher in a FRS while still being detectable by the face detector.

\noindent\textbf{Threat Model. }
We consider the scenario where the target FRS includes a ``\textit{no face alarm}" mechanism, which is triggered when the input image lacks a detectable face. This feature is commonly employed in security systems to ensure that the input image contains a visible and properly oriented face for accurate recognition. We present this approach for crafting a novel attack vector that bypass the ``no face alarm" in such systems through the use of adversarial images. This method takes into account the size constraint on the adversarial patterns, which is a crucial factor in determining the success of the attack.

We demonstrate that it is possible to craft powerful impersonation attacks that remain detectable to the face detection module of the target FRS, while still successfully executing the attack against the face matcher. Our work highlights the importance of considering patch size in adversarial patch design and offers a valuable tool for researchers and practitioners seeking to develop robust and efficient attacks against FRS with ``no face alarms".

\noindent\textbf{Attack fabrication. }
DE-Attack also leverage PGD method \cite{pgd} for crafting adversarial noise. The first adversarial objective of the DE-Attacks which is attacking MTCNN ($\mathcal{L}_{det-D}$), is the same as that of DI-Attacks as in Eq. \ref{det-D loss}.

\subsubsection{Evading Face Matcher}
The DE-Attack's second adversarial objective is to attack the feature extraction stage, resulting in evasion from true identities of the inputs. The evasion loss $\mathcal{L}_{eva}^{i}$ for the DE-Attack in the $i^{th}$ iteration of the attack generation process is calculated as follows

\begin{equation}\label{eq:evasion_face_matcher}
\mathcal{L}_{\text {eva}}^i=-\left.\left\|f\left(s_r\right)-f\left(s^i\right)\right\|\right|_p,
\end{equation}

\noindent where $s_r$ is the adversary's registered image with true identity, $s^i$ is the source image that adversary crafts to fool target FRS.

\subsubsection{Calculating Total Loss}
The total is then calculated as
\begin{equation}
\mathcal{L}_{\text {total}}^i=\alpha \cdot \mathcal{L}_{\text {det-D }}^i+\mathcal{L}_{\text {eva }}^i.
\end{equation}

The total loss $\mathcal{L}_{total}$ is finally optimized following the DI-Attack's methodology.

\subsection{UE-Attack}
The UE-Attack (Undetectable-Evasion Attack) method involves crafting evasion AXs that are designed to simultaneously evade both the face detector and face matcher in a FRS. 

\noindent \textbf{Threat Model. }
Mostly the systems like surveillance systems where the proper presentation and visibility of the face is not a necessary condition, do not have the ``no face alert" mechanism. For those kinds of systems, the adversarial objective can be achieved by evading either the face detector or the face matcher. Conventional attacks focus on either components independently, but it is possible to generate a much more robust evasion attack against the FRS as a whole that is crafted to fool both the face detector and face matcher. 

To ensure evasion attack robustness, even if it fails to fool a single component, our UE-Attack optimizes the adversarial noise to jointly evade the target FRS's face detector and face matcher, resulting in much more powerful adversarial attacks compared to conventional attacks. Additionally, our method allows for the crafting of adversarial patches of much smaller sizes than conventional methods, which successfully cause evasion against face detectors.

\noindent\textbf{Attack fabrication. }
UE-Attack's are also fabricated in the white-box PGD attack setting. Differently from DI and DE-Attacks, their attack against detection stage also targets to cause evasion.
However, the adversarial loss $\mathcal{L}^{i}_{eva}$ to cause the evasion against the face matcher stays the same as in Eq. \ref{eq:evasion_face_matcher}.
\subsubsection{Evading MTCNN}
The detection loss $\mathcal{L}_{det-E}$ in the $i^{th}$ iteration of the attack generation process for evading MTCNN is calculated by penalizing the probability $Y_{k,l}^{i}$ of a face in the input image with the adversarial patch, i.e.
\begin{equation}\label{det-E loss}
    \mathcal{L}_{det-E}^{i} = \sum_{k \in 1,2..m} \sum_{l \in 1,2..n} (A_{k,l}\cdot Y_{k,l}^{i} - K \cdot A_{k,l})^{s},
\end{equation}

\noindent where the variables has the same definition as in Eq. \ref{det-D loss}.

\subsubsection{Calculating Total Loss}
The total adversarial loss for the PGD attack objective is then calculated as 
\begin{equation}
\mathcal{L}_{\text {total}}^i=\alpha \cdot \mathcal{L}_{\text {det-E}}^i+\mathcal{L}_{\text {eva }}^i.
\end{equation}

Upon completion of the PGD optimization, the crafted attacks effectively evade the face detection as well as the face feature extraction and matching stages of the target FRS for the UE-Attack.

\section{Experiments}
In our experiments, we selected \textit{five} identities, with five images for each identity, from the \textit{VggFace2 dataset} \cite{cao2018vggface2} for each type of attack and employed the \textit{MTCNN} \cite{zhang2016joint} face detector and the \textit{Squeeze and Excitation-InceptionResNet50 (SEIR50)} \cite{hu2018squeeze} as the face feature extractor, which was trained on VggFace2 data. The test accuracy of the SEIR50 model for the test VggFace2 data was 99.37\%, showing good quality of the used trained model. We use adversarial patch noise in the form of \textit{eyeglass frames} in \textit{two different sizes}. The small patch is used to evaluate the evasion potential against the face detector, as the small size is not sufficient to hide representative facial features. The large frame hides these features and makes the face undetectable. We use the large eyeglass frame to evaluate our method in scenarios where the face must remain detectable to the face detector. The target identities for the impersonation attacks were also chosen from the VggFace2 dataset. We used the $L^2$-norm as the similarity measure. The attack method we used was the \textit{Projected Gradient Descent (PGD)} attack \cite{pgd}, and the adversarial objective was defined based on the type of generated attack. However, for the generation of the adversarial patches, we use PGD attack's $L^{\infty}$ constraint to the same values as the range of pixel values of the original images being crafted as AXs. All attacks were generated in the white-box setting, assuming the adversary has access to the full pipeline of the target FRS, and repeated \textit{thrice} to take into account the randomness during attack optimization.

\section{Results and Discussion}
\begin{table}[t]
\caption{Results for DI-Attacks}
\begin{center}
\begin{tabular}{|c|c|c|c|c|c|}
\hline
\textbf{Source}& \multicolumn{3}{|c|}{Mean Detection Probability (PNet)} & Mean & \multirow{2}{*}{OASR} \\ \cline{2-4}
ID & CI & CI+CP & CI+AP & ISR & \\
\hline
$S_1$ & 0.83 & $\textbf{0.54}\pm 0.018$ & \textbf{0.87} $\pm$ 0.030 & \textbf{1} & \textbf{1} \\
\hline
$S_2$ & 0.78 & $\textbf{0.49} \pm 0.042$ & \textbf{0.85} $\pm$ 0.280 & \textbf{1} & \textbf{0.67}\\
\hline
$S_3$ & 0.82 & $\textbf{0.51} \pm 0.002$ & \textbf{0.87} $\pm$ 0.057 & \textbf{1} & \textbf{1}\\
\hline
$S_4$ & 0.93 & $\textbf{0.55} \pm 0.029$ & \textbf{0.95} $\pm$ 0.140 & \textbf{1} & \textbf{1}\\
\hline
$S_5$ & 0.78 & $\textbf{0.47} \pm 0.031$ & \textbf{0.91} $\pm$ 0.086 & \textbf{1} & \textbf{1}\\
\hline
\multicolumn{6}{l}{CI: Clean Image; CP: Clean Patch; AP: Adversarial Patch}\\
\multicolumn{6}{l}{ISR: Impersonation Success Rate against face matcher only}\\
\multicolumn{6}{l}{OASR: Overall Attack Success Rate against the system as a whole}
\end{tabular}
\label{tab1}
\end{center}
\end{table}

\begin{table}[t]
\caption{Results for DE-Attacks}
\begin{center}
\begin{tabular}{|c|c|c|c|c|c|}
\hline
\textbf{Source}& \multicolumn{3}{|c|}{Mean Detection Probability (PNet)} & Mean & \multirow{2}{*}{OASR} \\ \cline{2-4}
ID & CI & CI+CP & CI+AP & ESR & \\
\hline
$S_1$ & 0.71 & $\textbf{0.48}\pm 0.033$ & \textbf{0.79} $\pm$ 0.054 & \textbf{1} & \textbf{1} \\
\hline
$S_2$ & 0.81 & $\textbf{0.52} \pm 0.007$ & \textbf{0.89} $\pm$ 0.027 & \textbf{1} & \textbf{1}\\
\hline
$S_3$ & 0.77 & $\textbf{0.47} \pm 0.112$ & \textbf{0.81} $\pm$ 0.011 & \textbf{1} & \textbf{1}\\
\hline
$S_4$ & 0.89 & $\textbf{0.51} \pm 0.078$ & \textbf{0.87} $\pm$ 0.120 & \textbf{1} & \textbf{1}\\
\hline
$S_5$ & 0.83 & $\textbf{0.54} \pm 0.045$ & \textbf{0.94} $\pm$ 0.080 & \textbf{1} & \textbf{1}\\
\hline
\multicolumn{6}{l}{CI: Clean Image; CP: Clean Patch; AP: Adversarial Patch}\\
\multicolumn{6}{l}{ESR: Evasion Success Rate against face matcher only}\\
\multicolumn{6}{l}{OASR: Overall Attack Success Rate against the system as a whole}
\end{tabular}
\label{tab2}
\end{center}
\end{table}

Table \ref{tab1} presents the results of \textit{DI-Attacks} on five different source identities (S1-S5). The mean detection probability (PNet) is reported for three different input types: clean images (CI), clean images with random patch noise (CI+CP), and clean images with adversarial patch noise (CI+AP). The mean impersonation success rate (ISR) demonstrates the attack precision against the face matcher, regardless of its performance against the detection stage. The overall ASR represents the performance of the generated attacks against both the face detection and matching phases. The results show that the mean detection probability decreases at first and the face becomes undetectable to the face detector of the target FRS due to the introduction of larger random adversarial patch noise in the form of eyeglass frames. However, after adversarial optimization of the noise for the joint attack objective, the clean image with eyeglass frame noise becomes detectable to the face detector. We found that our DI-Attack successfully made all input images from all identities detectable to the face detector. Finally, after calculating the mean ISR against the face matcher, we conclude that our attack could achieve a 100\% \textit{overall ASR} for all source identities except for S2, which has an ASR of 0.67, against the target FRS as a whole, demonstrating the effectiveness of our method.

Table \ref{tab2} presents the results for \textit{DE-Attacks}. It can be observed that the adversarial objective against both the face detection and face matching phases is achieved for all identities. The attacker was able to successfully make the face images with large eyeglass frame noise detectable to the face detector while also successfully causing evasion against the face matcher. Table \ref{tab3} presents the results for \textit{UE-Attacks}. In contrast to the DI and DE-Attacks, UE-Attacks aim to cause evasion against the face detector as well. Therefore, adding a large eyeglass frame noise that is not optimized to evade the face detector can easily evade the face matcher as well. However, to make the attack generation more challenging and to demonstrate that our method can create evasion against the face detection with very small, physically imperceptible patches, we used very thin eyeglass frame noise that is insufficient to hide representative facial features from the face detector, thus not causing evasion naturally. We found that the thin eyeglass frame noise, when converted to adversarial noise for our UE-Attack's joint adversarial objective against both the face detection and matching, resulted in successful evasion against both the face detector and face matcher for almost all identities, indicating the robustness of the attacks generated through our UE-Attack.

\begin{figure}[t]
\centerline{\includegraphics[width=0.91\linewidth]{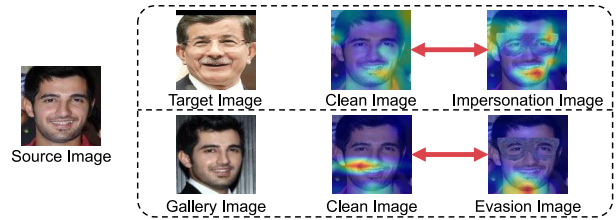}}
\caption{Understanding the decision-making process of a face matcher for clean, impersonation, and evasion adversarial images through Grad-CAM visualization \cite{selvaraju2017grad}. The top row illustrates the shift in representative pixels towards the adversarial eyeglass frame region in the impersonation adversarial image, leading to an impersonation error by the model. The bottom row shows that the adversarial eyeglass noise in the evasion attacks effectively hides the face from the model, causing the model to focus on random non-facial regions in the image. Face images are from the VggFace2 dataset \cite{cao2018vggface2}.}
\label{fig}
\end{figure}

\begin{table}[t]
\caption{Results for UE-Attacks}
\begin{center}
\begin{tabular}{|c|c|c|c|c|c|}
\hline
\textbf{Source}& \multicolumn{3}{|c|}{Mean Detection Probability (PNet)} & Mean & \multirow{2}{*}{OASR} \\ \cline{2-4}
ID & CI & CI+CP & CI+AP & ESR & \\
\hline
$S_1$ & 0.74 & $\textbf{0.68}\pm 0.033$ & \textbf{0.42} $\pm$ 0.072 & \textbf{1} & \textbf{1} \\
\hline
$S_2$ & 0.78 & $\textbf{0.72} \pm 0.052$ & \textbf{0.45} $\pm$ 0.098 & \textbf{1} & \textbf{1}\\
\hline
$S_3$ & 0.81 & $\textbf{0.66} \pm 0.067$ & \textbf{0.39} $\pm$ 0.189 & \textbf{1} & \textbf{1}\\
\hline
$S_4$ & 0.82 & $\textbf{0.82} \pm 0.041$ & \textbf{0.55} $\pm$ 0.043 & \textbf{1} & \textbf{1}\\
\hline
$S_5$ & 0.87 & $\textbf{0.77} \pm 0.032$ & \textbf{0.47} $\pm$ 0.113 & \textbf{1} & \textbf{1}\\
\hline
\multicolumn{6}{l}{CI: Clean Image; CP: Clean Patch; AP: Adversarial Patch}\\
\multicolumn{6}{l}{ESR: Evasion Success Rate against face matcher only}\\
\multicolumn{6}{l}{OASR: Overall Attack Success Rate against the system as a whole}
\end{tabular}
\label{tab3}
\end{center}
\end{table}

\subsection{Performance of Vanilla Attack Baselines}
In order to evaluate the effectiveness of traditional vanilla attacks on the face detection and matching system, we conducted experiments in which we focused on either the face detector or the face matcher as the target component. We generated both white-box evasion and impersonation attacks for both the detector and matcher, using two different patch sizes in the PGD attack setting. Our results showed that we were able to achieve 100\% ASRs against the target single model for both evasion and impersonation attacks. However, the transferability of these crafted attacks to the other component in the FRS was below 50\%, resulting in a below 50\% overall ASR for such attacks. This suggests that our attack methods provide strong attack robustness for simultaneous fooling of multiple components in the facial recognition system.

\subsection{Robustness Enhancement of Practical Face Recognition}
The primary importance of our proposed attacks lies in their ability to augment the adversarial robustness of practical FRSs. Our attacks provide more selective control during the robustness augmentation process for FRSs against novel practical attack vectors. These attacks are architectural agnostic, meaning they can be used to train \cite{advtraining} face detectors and feature extractors in any FRS pipeline. Additionally, the same set of adversarial data can be used to train multiple components within an FRS, providing space complexity benefits.

Our DE-Attack and DI-Attack, which cause fake face-presence-confidences to face detectors, can be used to train face detectors to discard such instances. On the other hand, the UE-Attack with small adversarial eyeglass frames can be used to train face detectors to correctly identify them. Adversarial detectors\cite{advdetectors}\cite{tramer2022detecting} can also be trained on these attacks and integrated into the target FRS to enhance robustness. Additionally, a simple input preprocessing methodology can be devised using information from our attacks to discard AXs. Overall, we provide valuable tools for improving the robustness of FRSs against AXs.

\section{Conclusion}
In this paper, we proposed multi-objective adversarial attacks on FRSs to evaluate the robustness of these systems against AXs. Our attacks effectively deceived both the face detection and face matching modules within an FRS, achieving up to 100\% ASRs and manipulating face detection probability by 50\%. This work analyzes the vulnerabilities of practical FRSs to sophisticated multi-objective adversarial attacks. Our attacks can be used to improve the robustness of these systems by leveraging them during training. Furthermore, the multi-objectiveness of our attacks provides space complexity benefits in the case of large datasets, as the same data can be used in training multiple models.

{\small
\bibliographystyle{plain}
\bibliography{egbib}
}

\end{document}